\documentclass[sigconf]{acmart}

\AtBeginDocument{%
  }

\setcopyright{acmlicensed}
\copyrightyear{2018}
\acmYear{2018}
\acmDOI{XXXXXXX.XXXXXXX}

\acmConference[Conference acronym 'XX]{Make sure to enter the correct
  conference title from your rights confirmation emai}{June 03--05,
  2018}{Woodstock, NY}
\acmISBN{978-1-4503-XXXX-X/18/06}

\usepackage{graphicx}
\usepackage{xspace}

\begin{document}

\title{Can Large Language Model Predict Employee Attrition?}


\author{Xiaoye Ma}
\affiliation{%
  \institution{Tomsk State University}
  \city{Tomsk}
  \country{Russia}
}\email{471019225@qq.com}

\author{Weiheng Liu}
\affiliation{%
  \institution{Tomsk State University}
  \city{Tomsk}
  \country{Russia}}
\email{1924262777@qq.com}

\author{Changyi Zhao}
\affiliation{%
  \institution{Tomsk State University}
  \city{Tomsk}
  \country{Russia}}
\email{zhaochangyi0120@gmail.com}

\author{Liliya R. Tukhvatulina}
\authornote{Corresponding author.}
\affiliation{%
  \institution{Tomsk State University}
  \city{Tomsk}
  \country{Russia}}
\email{ltuhvatulina@tpu.ru}

\renewcommand{\shortauthors}{Ma et al.}

\begin{abstract}
Employee attrition is a critical issue faced by organizations, with significant costs associated with turnover and the loss of valuable talent. Traditional methods for predicting attrition often rely on statistical techniques that, while useful, struggle to capture the complexity of modern workforces. Recent advancements in machine learning (ML) have provided more accurate, scalable solutions, allowing organizations to analyze diverse data points and predict attrition with greater precision. However, the emergence of large language models (LLMs) has opened new possibilities in human resource management by offering the ability to interpret contextual information from employee communications and detect subtle cues related to turnover.

In this paper, we leverage the IBM HR Analytics Employee Attrition dataset to evaluate the effectiveness of a fine-tuned GPT-3.5 model in comparison to traditional machine learning classifiers, including Logistic Regression, k-Nearest Neighbors (KNN), Support Vector Machine (SVM), Decision Tree, Random Forest, AdaBoost, and XGBoost. Our study focuses on assessing the predictive power, interpretability, and real-world applicability of each model. While traditional models offer ease of use and transparency, LLMs have the potential to uncover more nuanced patterns in employee behavior. Through our analysis, we aim to provide practical insights for organizations seeking to enhance their employee retention strategies with advanced predictive tools.

Our results show that the fine-tuned GPT-3.5 large language model (LLM) outperforms traditional machine learning approaches in terms of prediction accuracy, with an impressive precision of 0.91, recall of 0.94, and an F1-score of 0.92. In contrast, the best-performing traditional model, Support Vector Machine (SVM), achieved an F1-score of 0.82, while ensemble methods like Random Forest and XGBoost reached F1-scores of 0.80. These findings highlight the ability of GPT-3.5 to capture complex patterns in employee behavior and attrition risks, offering enhanced interpretability by identifying subtle linguistic cues and recurring themes. This demonstrates the potential of integrating LLMs into HR strategies to significantly improve predictive performance and decision-making in employee retention.

\end{abstract}

\ccsdesc[500]{Computing methodologies}
\ccsdesc[300]{Computing methodologies~Machine learning}

\keywords{Business, LLM, Machine learning}


\maketitle



\section{Introduction}

Employees are invaluable assets and investments for companies of all sizes, from small businesses to large multinational corporations. For instance, the average cost of a single interview in Switzerland is CHF 400, and the median cost to hire a skilled employee can reach CHF 9,738~\cite{blatter2016hiring}. While recruitment processes are designed to select the most suitable candidates, they do not always guarantee successful outcomes. Employee attrition has long been a challenge for human resource managers and employers, particularly in fast-growing economies~\cite{iqbal2010employee, alao2013analyzing}.

Employee attrition is a multifaceted issue influenced by various factors that impact organizational performance. Common causes include the absence of career growth opportunities, inadequate compensation, and poor work-life balance, all of which contribute to dissatisfaction and burnout~\cite{latha2013study, alao2013analyzing}. Additionally, mismanagement, a mismatch between job responsibilities and employee skills, and cultural disconnects within the organization can further exacerbate turnover~\cite{pallathadka2022attrition}. Employees may also leave due to better job offers, lack of recognition, unresolved workplace conflicts, or perceived job insecurity~\cite{raza2022predicting,burgard2009perceived}. Understanding these factors is crucial for organizations aiming to improve employee retention strategies.

Traditional approaches to predicting employee attrition typically involve the analysis of historical data using statistical methods to identify patterns and trends. These methods often require manual data analysis, such as reviewing exit interviews, employee surveys, and demographic information to detect common reasons for leaving. HR teams tend to focus on key indicators such as job tenure, performance reviews, and salary growth to assess the risk of turnover. However, these methods can be limited by their reliance on retrospective data and may struggle to capture the complexities of attrition dynamics in larger, more diverse workforces. They may also face challenges related to scalability and accuracy. Despite these limitations, traditional methods have served as foundational tools for understanding workforce trends and informing retention strategies before the advent of more advanced machine learning techniques.

To address these challenges, machine learning (ML) has gained traction as a more powerful method for employee attrition prediction. ML models can analyze vast amounts of data, uncover complex patterns, and make more accurate predictions than traditional approaches. By incorporating diverse data points, such as employee engagement levels, work performance, demographic characteristics, and external factors like market trends, ML models offer a more comprehensive view of attrition risks. These models—including classification algorithms like Logistic Regression, Decision Trees, Random Forests, and advanced techniques like XGBoost and neural networks—can continuously learn from new data, improving predictive performance over time.

Moreover, ML-based systems can handle large-scale, real-time data, making them highly adaptable to dynamic workforce environments. These systems can also detect early warning signs of attrition, enabling organizations to intervene proactively. By leveraging these predictive capabilities, businesses can take targeted actions, such as personalizing employee development programs, adjusting compensation structures, or addressing disengagement issues. In this way, ML offers a forward-looking, data-driven solution that empowers organizations to mitigate employee turnover and maintain a competitive edge in talent retention.

Recently, large language models (LLMs) have emerged as powerful tools in diverse fields, from natural language processing~\cite{gpt4-paper, gemini-paper, llama-paper} to scientific research~\cite{chen2023lm4hpc, ding2023hpc, chen2023data}, and their applications are now expanding into human resource management. LLMs enable organizations to go beyond numerical data and tap into the contextual richness present in employee communications. These models can analyze subtle linguistic patterns, shifts in tone, or recurring themes in employee communications to detect potential attrition risks. This deeper understanding of employee sentiment allows HR professionals to make more informed decisions about retention strategies. As LLMs continue to advance, their integration into employee attrition prediction models offers the potential to significantly enhance accuracy, scalability, and interpretability, thereby transforming workforce retention practices.

Despite the advantages of LLMs, their effectiveness in employee attrition prediction is still under scrutiny. In this work, we use the widely recognized IBM HR Analytics Employee Attrition dataset to compare the performance of a fine-tuned GPT-3.5 model with classic machine learning models, including Logistic Regression, k-Nearest Neighbors (KNN), Support Vector Machine (SVM), Decision Tree, Random Forest, AdaBoost, and XGBoost.

Our objective is to evaluate the predictive capabilities of the LLM against these traditional classifiers, with a particular focus on accuracy, interpretability, and real-world applicability. While traditional models have long been favored for attrition prediction due to their simplicity and interpretability, LLMs have the potential to uncover more complex patterns in employee behavior. This study aims to determine whether fine-tuned LLMs can outperform or complement traditional methods and provide unique insights into the drivers of employee turnover.

By evaluating the strengths and weaknesses of each model, we aim to offer practical guidance for organizations seeking to integrate advanced predictive tools into their human resource strategies. Additionally, we explore how model interpretability—crucial in HR decisions—differs across approaches, particularly in terms of transparency and decision-making between LLMs and more conventional machine learning methods.

The remainder of this paper is organized as follows. Section 2 provides an overview of existing research on employee attrition prediction, focusing on both traditional machine learning models and recent applications of large language models (LLMs). We discuss the strengths and limitations of various approaches and highlight the research gap that this study aims to address. Section 3 describes the IBM HR Analytics Employee Attrition dataset used in our analysis. We outline the key attributes of the dataset, including employee demographics, job satisfaction, and performance indicators. We also discuss the preprocessing steps, such as data cleaning, handling missing values, encoding categorical features, and normalizing the data to prepare it for model training. Section 4 details the methodology used to train and evaluate both the traditional machine learning models and the fine-tuned GPT-3.5 model. We explain the parameter tuning process, feature selection, and model training strategies. Additionally, we describe the evaluation metrics used to compare model performance, such as accuracy, precision, recall, and F1-score, as well as considerations for model interpretability. Section 5 presents the results of our experiments, comparing the performance of each model on the employee attrition prediction task. We provide a detailed analysis of the strengths and weaknesses of both traditional machine learning models and the LLM in terms of predictive accuracy, interpretability, and practical applicability in HR decision-making.

The main contributions of this paper are as follows:
\begin{itemize}
    \item detailed steps in processing the IBM HR Analytics Employee Attrition dataset for ML and LLM training.
    \item the pioneering effort to fine-tune GPT model for the task of employee attrition.
    \item practical insights for ML approach in HR Management.
\end{itemize}

\section{Background}
This section provides background information on employee attrition and large language models.
\subsection{Employee Attrition}
Employee attrition refers to the voluntary or involuntary departure of employees from an organization, and it is a significant concern for businesses due to its impact on operational efficiency and costs. High attrition rates can disrupt productivity, increase the financial burden of hiring and training new employees, and negatively affect overall employee morale. Attrition can be categorized into two main types: voluntary attrition, where employees leave by their own choice (e.g., for a better opportunity or dissatisfaction with the current role), and involuntary attrition, where the employer terminates the employee’s contract due to factors like poor performance or organizational restructuring.

Several factors contribute to employee attrition. These may include personal reasons such as lack of career development opportunities, inadequate compensation, poor work-life balance, and limited recognition for accomplishments. Organizational factors, such as unsatisfactory leadership, misalignment between job responsibilities and employee skills, toxic workplace culture, and perceived job insecurity, are also major drivers of turnover~\cite{pallathadka2022attrition}. External factors, including changes in the job market or better offers from competing firms, further exacerbate this issue~\cite{raza2022predicting}.

For organizations, the consequences of attrition extend beyond the cost of replacing employees. The hiring process itself is expensive and time-consuming, as it involves recruitment, interviews, onboarding, and training. Additionally, the loss of experienced employees can lead to a temporary decline in organizational knowledge and productivity, especially if the departing employees hold critical positions. According to a 2016 report by the Society for Human Resource Management, the average cost of hiring a new employee exceeds \$4,000, and the average annual turnover rate in the U.S. is estimated at 19\%~\cite{blatter2016hiring}. Therefore, predicting and preventing employee attrition has become a strategic priority for HR departments, prompting the development of more advanced analytical techniques to identify at-risk employees and implement targeted retention strategies.

\subsection{Employee Attrition Prediction Approach}

Traditional methods for predicting employee attrition involve manual analysis of historical data, such as exit interviews and performance reviews. However, these approaches are limited in scope, accuracy, and scalability, especially for large organizations with complex dynamics. Machine learning (ML) has emerged as a more advanced solution, offering higher accuracy and adaptability. ML algorithms can analyze large volumes of data, identifying complex, non-linear relationships between employee characteristics and turnover risks. Common ML techniques for attrition prediction include Logistic Regression, Decision Trees, Random Forests, and more advanced methods like Support Vector Machines and XGBoost. These models process various structured data points, including employee demographics, tenure, and performance evaluations, while also incorporating external factors such as market trends and economic conditions. This comprehensive approach allows organizations to build more robust predictive models. Several studies~\cite{raza2022predicting, alduayj2018predicting, jain2020explaining} have demonstrated the efficacy of different ML approaches in predicting employee attrition.

\subsection{Large Language Models}

Large Language Models (LLMs) have revolutionized Natural Language Processing (NLP) since the introduction of transformer-based models by Vaswani et al.~\cite{vaswani2017attention} in 2017. These models employ self-attention mechanisms to effectively capture long-range dependencies in text, outperforming traditional recurrent neural networks. The multi-head attention architecture enables parallel processing of multiple attention patterns, addressing issues like vanishing gradients and improving scalability. LLMs, built on transformer principles, are trained on vast datasets to model complex language distributions. Their exceptional performance in NLP~\cite{gpt4-paper,chen2024landscape} and scientific applications~\cite{chen2024ompgpt,chen2022multi} has led to increased exploration of their potential in human resource management, including applications in management consulting~\cite{mohan2024management}, HR chatbots~\cite{singh2023exploring}, and employee development~\cite{budhwar2023human}.

\section{Dataset Prepare}
This section introduces the dataset used in this work. We performed different analyses to present the data from different perspectives and detailed our methods to process the data for machine learning model training and testing.

\subsection{Dataset Introducation}

The IBM HR Analytics Employee Attrition \& Performance dataset is a well-structured collection of employee data aimed at identifying factors contributing to employee attrition. It contains detailed information on 1,470 employees, with 35 different attributes that offer a comprehensive overview of various aspects related to employee performance, demographics, and job satisfaction. This dataset provides a valuable foundation for HR analytics and predictive modeling, making it suitable for machine learning tasks focused on employee attrition prediction. The dataset has below basic characteristics:
\begin{itemize}
    \item Size: The dataset consists of 1,470 rows, each representing a unique employee, and 35 columns, each capturing specific attributes related to employee demographics, job roles, and performance metrics.
    \item Format: The dataset is provided in CSV format, making it easily accessible and compatible with most data processing tools and machine learning libraries.
    \item Missing Values: The dataset contains no missing values, with all columns having 1,470 non-null entries. This ensures data integrity and eliminates the need for extensive data cleaning, allowing for a more straightforward preprocessing pipeline.
\end{itemize}



\subsection{Key Features}
\begin{table*}[ht]
\centering
\caption{Features in the IBM HR Analytics Employee Attrition \& Performance Dataset}

\begin{tabular}{|p{0.3\textwidth}|p{0.6\textwidth}|}
\hline
\textbf{Category} & \textbf{Features} \\
\hline
Demographic Information & Age, Gender, Marital Status, Education, Education Field \\
\hline
Job-related Attributes & Department, Job Role, Job Level, Job Involvement, Job Satisfaction, Years at Company, Years in Current Role, Years Since Last Promotion, Years with Current Manager \\
\hline
Compensation and Benefits & Daily Rate, Hourly Rate, Monthly Income, Monthly Rate, Percent Salary Hike, Stock Option Level \\
\hline
Performance Metrics & Performance Rating, Training Times Last Year \\
\hline
Work Environment Factors & Business Travel, Environment Satisfaction, Work-Life Balance, Overtime \\
\hline
Other Relevant Information & Distance from Home, Relationship Satisfaction, Total Working Years, Number of Companies Worked \\
\hline
Target Variable & Attrition (Yes/No) \\
\hline
\end{tabular}
\label{tab:hr_analytics_features}
\end{table*}

The IBM HR Analytics Employee Attrition \& Performance dataset, as summarized in Table~\ref{tab:hr_analytics_features}, provides a comprehensive set of features for analyzing factors contributing to employee attrition. The dataset encompasses a wide array of information, from basic demographic data (such as age, gender, and marital status) to detailed job-related attributes (including department, job role, and tenure metrics). Additionally, it contains data on compensation, benefits, performance metrics, and work environment factors. Notably, features related to employee satisfaction, such as job satisfaction, work-life balance, and relationship satisfaction, offer insights into the non-monetary aspects of employee experience. Other variables like distance from home and the number of companies previously worked for may highlight less obvious influences on attrition. With attrition as the binary target variable, this feature set enables robust analysis and predictive modeling, helping HR professionals better understand and manage employee turnover.

Table~\ref{tab:hr_analytics_stats} presents descriptive statistics for key numerical features in the dataset, which may influence employee attrition. Each feature is described by its count, mean, standard deviation, and percentiles (minimum, 25th, 50th, 75th, and maximum values).

For example, the average employee age in the dataset is approximately 36.92 years, with a standard deviation of 9.14 years, while monthly income has a mean of \$6,502.93, ranging from \$1,009 to \$19,999. These statistics provide a snapshot of the dataset's distribution and will serve as the foundation for our predictive modeling of employee attrition patterns.

\begin{table*}[h]
\centering
\small
\caption{Descriptive Statistics of Key Numerical Features in the HR Analytics Dataset}
\begin{tabular}{|l|r|r|r|r|r|r|r|r|}
\hline
\textbf{Feature} & \textbf{Count} & \textbf{Mean} & \textbf{Std} & \textbf{Min} & \textbf{25\%} & \textbf{50\%} & \textbf{75\%} & \textbf{Max} \\
\hline
Age & 1470 & 36.92 & 9.14 & 18 & 30 & 36 & 43 & 60 \\
Daily Rate & 1470 & 802.49 & 403.51 & 102 & 465 & 802 & 1157 & 1499 \\
Distance from Home & 1470 & 9.19 & 8.11 & 1 & 2 & 7 & 14 & 29 \\
Education & 1470 & 2.91 & 1.02 & 1 & 2 & 3 & 4 & 5 \\
Monthly Income & 1470 & 6502.93 & 4707.96 & 1009 & 2911 & 4919 & 8379 & 19999 \\
Number of Companies Worked & 1470 & 2.69 & 2.50 & 0 & 1 & 2 & 4 & 9 \\
Percent Salary Hike & 1470 & 15.21 & 3.66 & 11 & 12 & 14 & 18 & 25 \\
Total Working Years & 1470 & 11.28 & 7.78 & 0 & 6 & 10 & 15 & 40 \\
Training Times Last Year & 1470 & 2.80 & 1.29 & 0 & 2 & 3 & 3 & 6 \\
Years at Company & 1470 & 7.01 & 6.13 & 0 & 3 & 5 & 9 & 40 \\
Years in Current Role & 1470 & 4.23 & 3.62 & 0 & 2 & 3 & 7 & 18 \\
Years Since Last Promotion & 1470 & 2.19 & 3.22 & 0 & 0 & 1 & 3 & 15 \\
Years with Current Manager & 1470 & 4.12 & 3.57 & 0 & 2 & 3 & 7 & 17 \\
\hline
\end{tabular}
\label{tab:hr_analytics_stats}
\end{table*}

\subsection{Data Cleaning and Reduction}

To ensure the dataset is clean and relevant for our analysis, we performed several preprocessing steps. First, the dataset, which contains a variety of employee attributes, was loaded. Some of these attributes were redundant or not useful for predicting employee attrition.

We reduced the dataset by removing the following features:
\begin{itemize}
    \item \textbf{EmployeeCount}: This feature is a constant value for all entries and thus provides no useful information for predictive modeling.
    \item \textbf{StandardHours}: Similar to \textit{EmployeeCount}, this column contains the same value for every employee and does not contribute to the variability needed for classification tasks.
    \item \textbf{Over18}: This feature contains only a single value (``Yes'') across all instances, which does not provide any useful distinction between employees.
    \item \textbf{EmployeeNumber}: This is simply an identifier for each employee and does not influence the outcome of attrition prediction.
\end{itemize}

By removing these features, we eliminated unnecessary columns and reduced the dimensionality of the data, which can improve the model's performance by focusing only on the relevant attributes. This step is crucial for ensuring that the model is not overfitting on irrelevant or redundant data.

\subsection{Data Analysis}
\begin{figure}[h]
    \centering
    \includegraphics[width=0.4\textwidth]{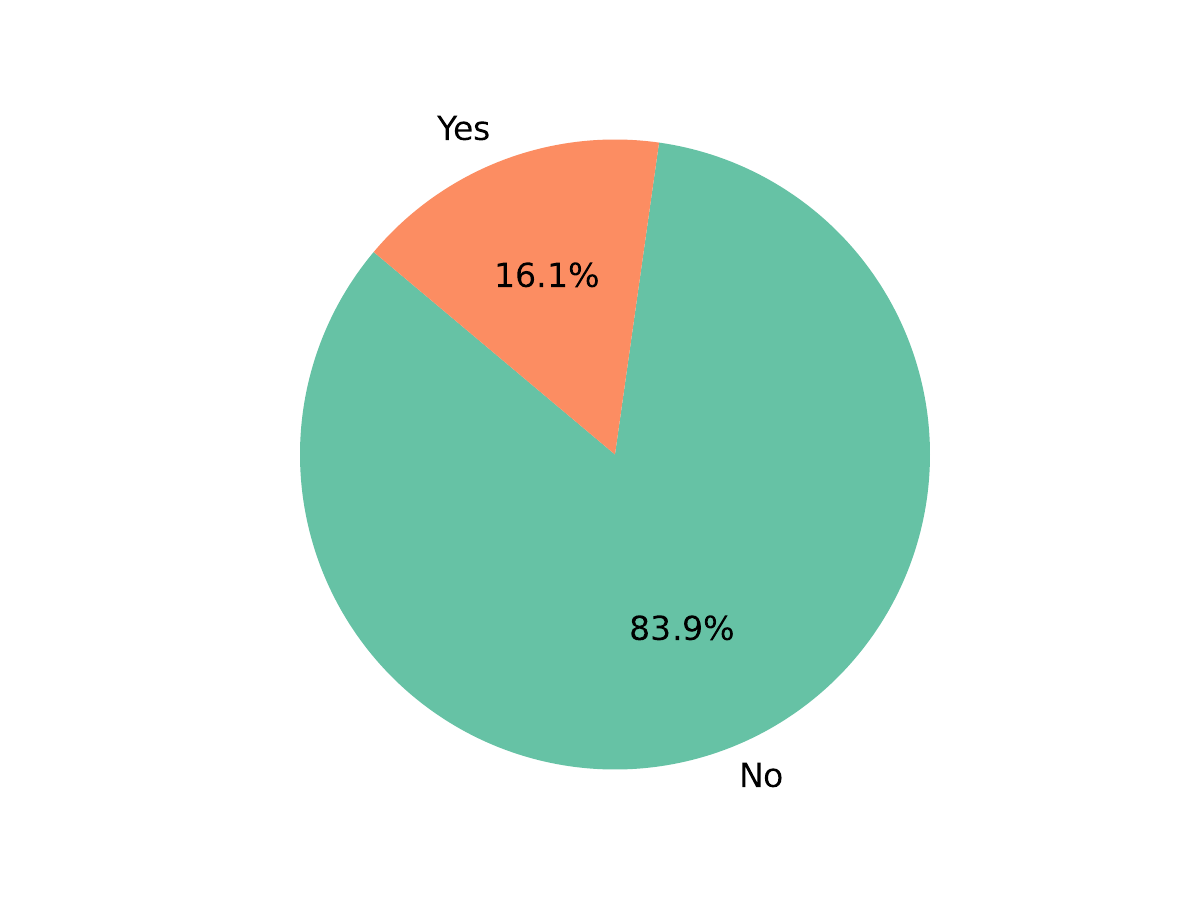}
    \caption{Employee Attrition Distribution}
    \label{fig:attrition_distribution}
\end{figure}

\textbf{Data Balance}. Figure~\ref{fig:attrition_distribution} illustrates the distribution of employee attrition within the dataset after cleaning and reduction. The majority of employees, 83.9\%, have not left the company, while only 16.1\% have. This substantial imbalance in the attrition classes highlights the need to address class imbalance when building predictive models. 

In particular, techniques such as oversampling the minority class or applying specialized algorithms for imbalanced data may be necessary to ensure that the model does not develop a bias towards predicting the majority class. Furthermore, the imbalance suggests that employee attrition is relatively uncommon, which could indicate either successful employee retention strategies or natural tendencies within the workforce.

\begin{figure*}[h]
    \centering
    \includegraphics[width=0.85\textwidth]{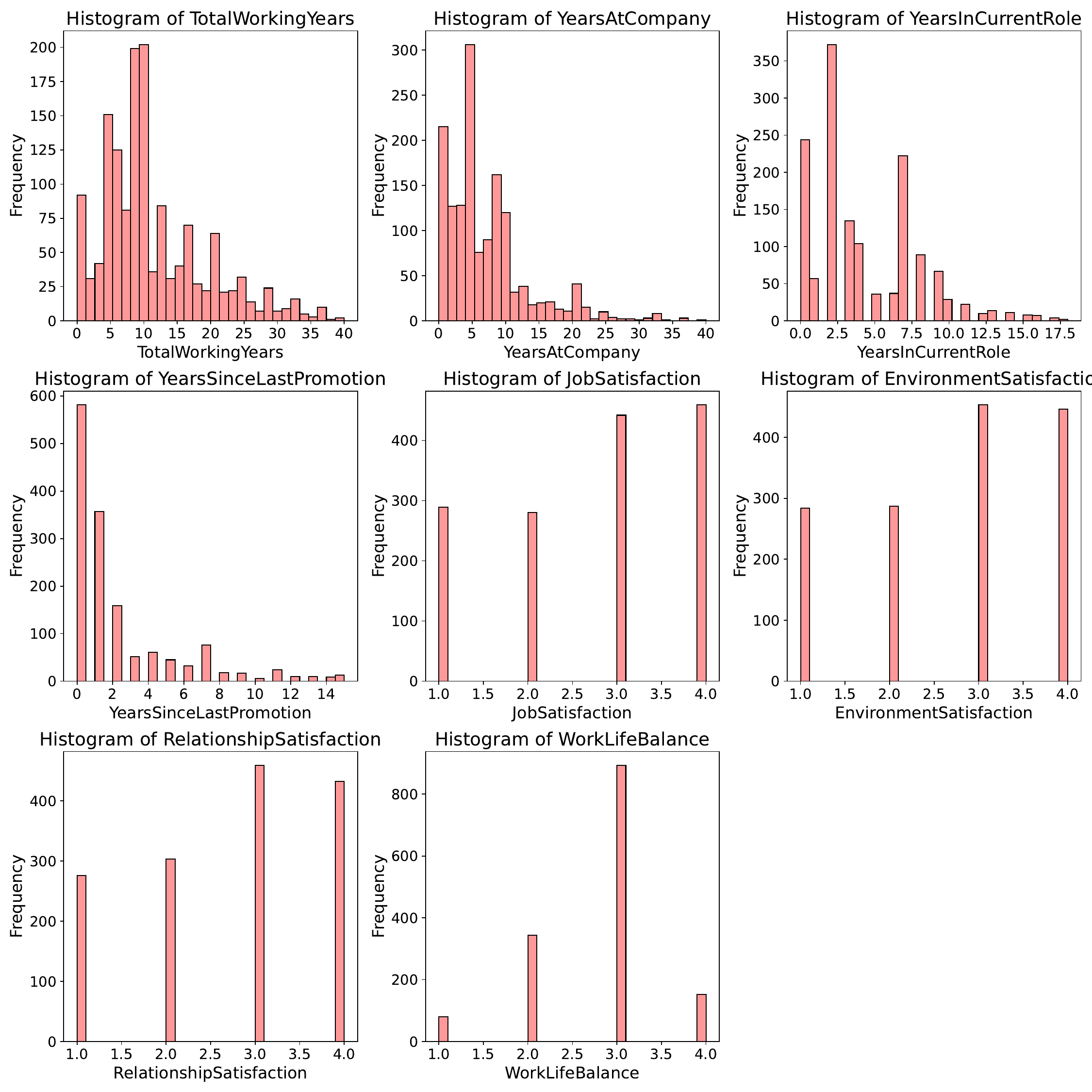}
    \caption{Frequency distribution of a feature by dividing the data into bins. The listed histograms show a skewed data issue in the corresponding feature.}
    \label{fig:Frequency_distribution}
\end{figure*}

\textbf{Frequency Distribution}.
To better understand the distribution of the numerical features in the dataset, we generated the histogram plot to provide insights into the data distributions. A histogram visualizes the frequency distribution of a feature by dividing the data into bins. The height of each bin represents the number of observations that fall within the corresponding range of values. This helps to identify skewness and the overall spread of the data. Figure~\ref{fig:Frequency_distribution} listed the features that have skewness issue.

\subsection{Skewed Data Handling}
Figure~\ref{fig:Frequency_distribution} illustrated features that have skewed issues. In our dataset, several numerical features exhibited \textit{right-skewed} distributions, where the majority of the data points are concentrated on the lower end of the scale, with a long tail extending to the right. Right-skewed distributions are common when certain features, such as income or years of experience, have a small number of very large values that disproportionately influence the distribution.

Skewed data can pose significant challenges in predictive modeling. Many machine learning algorithms, especially linear models and algorithms that assume normally distributed data, can be sensitive to skewness. Highly skewed features can result in models that are less accurate or biased toward the skewed data. By addressing skewness, we ensure that our models perform more effectively by learning from data that is better distributed, leading to more accurate predictions.

To address the issue of right-skewness, we applied a \textit{logarithmic transformation} to the affected features. For each numerical feature with a skewness greater than 0.5, we performed a log transformation using the $\log(1+x)$ function. This method helps reduce skewness by compressing large values and spreading out smaller ones, bringing the distribution closer to normal. After applying the transformation, the features exhibit a more symmetric distribution, improving the model’s ability to generalize and make accurate predictions.

\subsection{Feature Engineering}
To enhance the predictive power of the model and simplify the dataset, feature engineering was performed by creating two new composite features: \textit{WorkExperience} and \textit{OverallSatisfaction}. These features encapsulate important aspects of employees' work experience and satisfaction, reducing redundancy in the dataset while maintaining essential information for modeling.

The \textit{WorkExperience} feature was derived by averaging several key experience-related variables, including \textit{TotalWorkingYears}, \textit{YearsAtCompany}, \textit{YearsInCurrentRole}, \textit{YearsSinceLastPromotion}, and \textit{YearsWithCurrManager}. By consolidating these columns, a single metric representing an employee’s overall work experience was generated, simplifying the data without losing critical information.

Similarly, the \textit{OverallSatisfaction} feature was created by averaging the \textit{JobSatisfaction}, \textit{EnvironmentSatisfaction}, \textit{RelationshipSatisfaction}, and \textit{WorkLifeBalance} columns. This new feature provides a comprehensive measure of an employee’s satisfaction with both their job and work environment.

After generating these new features, the original columns used to create them were removed from the dataset. This step helped to reduce redundancy and improve the dataset’s efficiency, ensuring that the key information necessary for predictive modeling was retained in a more streamlined form.

\subsection{Categorical Encoding}

To prepare the categorical features for modeling, we applied label encoding to all categorical variables. Label encoding converts categorical values into numerical representations, which are required by most machine learning algorithms. Specifically, we identified the categorical columns in the dataset and used the \texttt{LabelEncoder} class from the \texttt{sklearn} library to encode each column. This process ensured that categorical features such as \textit{Department} and \textit{JobRole} were transformed into numerical values, enabling the machine learning model to process them effectively.

\subsection{Train-Test Splitting}

Once the data was prepared, we split it into training and test sets to evaluate the performance of the model. Using the \texttt{train\_test\_split} function from \texttt{sklearn}, we partitioned the dataset, reserving 80\% of the data for training and 20\% for testing. The target variable, \textit{Attrition}, was stratified to ensure that the class distribution in the training and test sets remained consistent. 

The resulting split produced 1176 samples for training and 294 samples for testing, as shown in the following output:
\begin{itemize}
    \item \textbf{X\_train shape:} (1176, 24)
    \item \textbf{X\_test shape:} (294, 24)
    \item \textbf{y\_train shape:} (1176,)
    \item \textbf{y\_test shape:} (294,)
\end{itemize}

This splitting ensures that the model can be trained on the larger portion of the data while being evaluated on a separate, unseen test set, providing an accurate assessment of the model's generalization performance.

\subsection{Feature Scaling}

Before training our machine learning model, it was essential to standardize the features to ensure that they are on the same scale. Standardization transforms features so that they have a mean of zero and a standard deviation of one. This process is particularly important for models that are sensitive to the magnitude of the input features, such as logistic regression and support vector machines. By standardizing the features, we ensure that the model does not place undue emphasis on features with larger ranges or units, which could negatively impact the performance of the model.

The training data was used to compute the scaling factors (mean and standard deviation), and these factors were applied consistently to both the training and test datasets. This ensures that the model generalizes well to unseen data while maintaining consistency across datasets.

\subsection{Imbalanced Data Handling}

Our dataset exhibited a significant class imbalance, with the number of non-attrition cases far outnumbering the attrition cases. Imbalanced datasets can lead to biased models that perform well on the majority class but poorly on the minority class, which in our case is employee attrition. To address this imbalance, we applied the Synthetic Minority Over-sampling Technique (SMOTE). 

SMOTE generates synthetic samples of the minority class by interpolating between existing samples, effectively increasing the number of minority class instances and balancing the class distribution. Before applying SMOTE, the training dataset contained 986 non-attrition samples and 190 attrition samples. After oversampling, the dataset was balanced, with 986 samples for each class. 

This balancing process ensures that the model has equal exposure to both classes, improving its ability to accurately detect instances of employee attrition and reducing bias toward the majority class. As a result, the model is better equipped to generalize to real-world scenarios where both classes are important.

\section{Experiment}
This section details the experiments, covering model selection and result comparisons.

\subsection{Classic Models Used for Attrition Prediction}

To build a robust and accurate predictive model for employee attrition, we employed a diverse set of machine learning algorithms. Each of these models offers unique strengths and approaches to classification, allowing us to compare their performance and identify the most suitable method for our dataset. The following models were used in our analysis:

\noindent
\textbf{Logistic Regression} estimates class probabilities using a logistic function on input features. This sklearn implementation uses default parameters: L2 regularization with C=1.0, 'lbfgs' solver, and no penalty. It offers interpretable predictions by revealing feature contributions. The model is trained on X\_train and y\_train, then used to predict both training and test data outcomes.

\noindent
\textbf{K-Nearest Neighbors (KNN)} is a non-parametric algorithm classifying instances based on proximity to neighbors in feature space. We used 10 neighbors, uniform weights, and Minkowski distance (p=2, equivalent to Euclidean). The algorithm choice is automatic, with a leaf size of 50 for optimized tree searches. KNN works well for non-linear boundaries in smaller datasets but is sensitive to distance metrics and k values.

\noindent
\textbf{Support Vector Machine (SVM)} finds the optimal hyperplane to separate classes in feature space. This implementation uses an RBF kernel with C=200, gamma='scale', and degree=5. It employs a 'one-vs-rest' decision function, tolerance of 0.1, and no probability estimates. SVM excels in high-dimensional spaces and can handle non-linear classification through kernel functions. It's particularly effective for datasets with clear class separation.

\noindent
\textbf{Decision Tree Classifier} creates a tree structure by splitting data based on feature value.  Decision Trees are easy to interpret and visualize but can be prone to overfitting, especially with small datasets. We limited tree depth to 5 levels, preventing overfitting.

\noindent
\textbf{Random Forest} combines multiple decision trees to improve accuracy and reduce overfitting.  Our implementation uses 200 trees, each with a maximum depth of 5. It employs 'sqrt' max\_features, selecting the square root of total features for each split. By averaging predictions from diverse trees, Random Forest handles complex feature relationships and outperforms single decision trees.

\noindent
\textbf{AdaBoost Classifier} combines weak classifiers to create a strong one, focusing on misclassified instances. Our implementation uses 150 estimators with a learning rate of 0.01. It employs the 'SAMME' algorithm for multiclass problems. AdaBoost iteratively adjusts instance weights, improving overall accuracy and reducing bias. It's effective for moderately complex datasets, balancing between weak classifiers to form a robust ensemble.

\noindent
\textbf{XGBoost} (Extreme Gradient Boosting) is an advanced gradient boosting implementation. Our configuration uses binary logistic regression for classification, with a learning rate of 0.01 and 350 estimators. Tree depth is limited to 3, and 'sqrt' max\_features is employed. XGBoost combines regularization and optimized computation, excelling in speed and accuracy for large, imbalanced datasets. Its flexibility allows for fine-tuning across various machine learning tasks.

\subsection{GPT Model for Attrition Prediction}

\textbf{GPT-3.5 Turbo} is a large language model that leverages the transformer architecture to perform predictive tasks. While it is predominantly used in natural language processing tasks, GPT-3.5 can be fine-tuned for various prediction and classification problems, including employee attrition prediction. We followed the process described in ~\cite{liga2023fine} to fine-tune GPT-3.5 turbo using prompt-response pairs constructed by the training set.

\noindent
\textbf{Data prepare.} We constructed a JSONL format data, in which each line is a prompt-response pair for the model to learn from.

\begin{verbatim}
{
    "prompt": "Analyze the employee information and 
    predict employee turnover: {Employee_Info}",
    "completion": "{Attrition_result}"
}
\end{verbatim}




\noindent
\textbf{GPT-3.5 Turbo Fine-tuning.} We used the OpenAI library and API to fine-tune the GPT model with our data. We set the $max\_token$ to be 50 to save the cost as our predicted results are $yes$ or $no$.

\noindent
\textbf{Inference.} To generate predictions, we passed the test set to the fine-tuned GPT model using its fine-tuned ID.

\subsection{Evaluation Metrics}
To evaluate the predictive models' performance, we used the classification metrics \textit{precision}, \textit{recall}, and \textit{F1-score}, as defined in Equations~\ref{eq:precision}, \ref{eq:recall}, and \ref{eq:f1}:

\begin{equation}
\text{Precision} = \frac{\text{True Positives}}{\text{True Positives} + \text{False Positives}}
\label{eq:precision}
\end{equation}

\begin{equation}
\text{Recall} = \frac{\text{True Positives}}{\text{True Positives} + \text{False Negatives}}
\label{eq:recall}
\end{equation}

\begin{equation}
\text{F1-Score} = 2 \cdot \frac{\text{Precision} \cdot \text{Recall}}{\text{Precision} + \text{Recall}}
\label{eq:f1}
\end{equation}
These metrics allow us to evaluate the trade-offs between precision and recall and help in determining the best performing model for employee attrition prediction.

\subsection{Results}

The comparative analysis of various machine learning models for this classification task revealed intriguing insights, as listed in Table~\ref{tab:model-comparison}. The fine-tuned GPT-3.5 model emerged as the superior performer, achieving exceptional metrics with a precision of 0.91, recall of 0.94, and an F1-score of 0.92. This performance significantly outpaced traditional machine learning algorithms, underscoring the potential of large language models in classification tasks when properly fine-tuned. Among conventional models, Support Vector Machine (SVM) demonstrated the strongest performance with an F1-score of 0.82, followed closely by ensemble methods such as Random Forest and XGBoost, both achieving an F1-score of 0.80. Logistic Regression and AdaBoost showed comparable efficacy with F1-scores of 0.78 and 0.79 respectively, while K-Nearest Neighbors (KNN) exhibited the lowest performance with an F1-score of 0.71. These results highlight the robust performance of ensemble methods and the particular suitability of SVM for this dataset. The marked superiority of the fine-tuned GPT-3.5 model suggests promising avenues for leveraging advanced language models in classification tasks, potentially revolutionizing approaches to similar problems in the field.

\begin{table}[h]
\centering
\caption{Comparison of Machine Learning Models (Weighted Average)}
\begin{tabular}{|l|c|c|c|}
\hline
\textbf{Model} & \textbf{Precision} & \textbf{Recall} & \textbf{F1-score} \\
\hline
Logistic Regression & 0.84 & 0.76 & 0.78 \\
KNN & 0.80 & 0.66 & 0.71 \\
SVM & 0.81 & 0.83 & 0.82 \\
Decision Tree & 0.83 & 0.78 & 0.80 \\
Random Forest & 0.82 & 0.80 & 0.80 \\
AdaBoost & 0.82 & 0.77 & 0.79 \\
XGBoost & 0.80 & 0.80 & 0.80 \\
GPT 3.5 Fine-tuned & \textbf{0.91} & \textbf{0.94} & \textbf{0.92} \\
\hline
\end{tabular}

\label{tab:model-comparison}
\end{table}

\section{Conclusion}

In this study, we explored the performance of various machine learning models for the task of employee attrition prediction, comparing traditional algorithms with the fine-tuned GPT-3.5 model. Our experiments demonstrated that the fine-tuned GPT-3.5 model significantly outperformed conventional machine learning techniques, achieving the highest precision, recall, and F1-score. This result underscores the potential of large language models in classification tasks, particularly when fine-tuned for domain-specific data.

Among traditional models, Support Vector Machine (SVM) exhibited the strongest performance, followed closely by ensemble methods such as Random Forest and XGBoost. These models also showed competitive results, validating their efficacy for this type of classification problem. However, simpler models like Logistic Regression and K-Nearest Neighbors (KNN) demonstrated lower predictive power, particularly in capturing the complexity of the data.

The superior performance of GPT-3.5 highlights the growing role that advanced language models can play in solving classification tasks beyond natural language processing. As machine learning techniques evolve, integrating large language models into predictive analytics pipelines holds immense potential for enhancing decision-making processes in fields like human resource management.

Future work could further explore the impact of different feature subsets on model performance for a comprehensive understanding of the factors influencing model performance.  Additionally, investigating the interpretability of these models in real-world applications will be crucial for their broader adoption in business and operational environments. Overall, this study demonstrates the advantages of leveraging LLMs in classification tasks and points to promising avenues for future research in this area.

\bibliographystyle{ACM-Reference-Format}
\bibliography{main}


\begin{thebibliography}{23}


\ifx \showCODEN    \undefined \def \showCODEN     #1{\unskip}     \fi
\ifx \showDOI      \undefined \def \showDOI       #1{#1}\fi
\ifx \showISBNx    \undefined \def \showISBNx     #1{\unskip}     \fi
\ifx \showISBNxiii \undefined \def \showISBNxiii  #1{\unskip}     \fi
\ifx \showISSN     \undefined \def \showISSN      #1{\unskip}     \fi
\ifx \showLCCN     \undefined \def \showLCCN      #1{\unskip}     \fi
\ifx \shownote     \undefined \def \shownote      #1{#1}          \fi
\ifx \showarticletitle \undefined \def \showarticletitle #1{#1}   \fi
\ifx \showURL      \undefined \def \showURL       {\relax}        \fi
\providecommand\bibfield[2]{#2}
\providecommand\bibinfo[2]{#2}
\providecommand\natexlab[1]{#1}
\providecommand\showeprint[2][]{arXiv:#2}

\bibitem[Achiam et~al\mbox{.}(2023)]%
        {gpt4-paper}
\bibfield{author}{\bibinfo{person}{Josh Achiam}, \bibinfo{person}{Steven Adler}, \bibinfo{person}{Sandhini Agarwal}, \bibinfo{person}{Lama Ahmad}, \bibinfo{person}{Ilge Akkaya}, \bibinfo{person}{Florencia~Leoni Aleman}, \bibinfo{person}{Diogo Almeida}, \bibinfo{person}{Janko Altenschmidt}, \bibinfo{person}{Sam Altman}, \bibinfo{person}{Shyamal Anadkat}, {et~al\mbox{.}}} \bibinfo{year}{2023}\natexlab{}.
\newblock \showarticletitle{Gpt-4 technical report}.
\newblock \bibinfo{journal}{\emph{arXiv preprint arXiv:2303.08774}} (\bibinfo{year}{2023}).
\newblock


\bibitem[Alao and Adeyemo(2013)]%
        {alao2013analyzing}
\bibfield{author}{\bibinfo{person}{DABA Alao} {and} \bibinfo{person}{AB Adeyemo}.} \bibinfo{year}{2013}\natexlab{}.
\newblock \showarticletitle{Analyzing employee attrition using decision tree algorithms}.
\newblock \bibinfo{journal}{\emph{Computing, Information Systems, Development Informatics and Allied Research Journal}} \bibinfo{volume}{4}, \bibinfo{number}{1} (\bibinfo{year}{2013}), \bibinfo{pages}{17--28}.
\newblock


\bibitem[Alduayj and Rajpoot(2018)]%
        {alduayj2018predicting}
\bibfield{author}{\bibinfo{person}{Sarah~S Alduayj} {and} \bibinfo{person}{Kashif Rajpoot}.} \bibinfo{year}{2018}\natexlab{}.
\newblock \showarticletitle{Predicting employee attrition using machine learning}. In \bibinfo{booktitle}{\emph{2018 international conference on innovations in information technology (iit)}}. IEEE, \bibinfo{pages}{93--98}.
\newblock


\bibitem[Blatter et~al\mbox{.}(2016)]%
        {blatter2016hiring}
\bibfield{author}{\bibinfo{person}{Marc Blatter}, \bibinfo{person}{Samuel Muehlemann}, \bibinfo{person}{Samuel Schenker}, {and} \bibinfo{person}{Stefan~C Wolter}.} \bibinfo{year}{2016}\natexlab{}.
\newblock \showarticletitle{Hiring costs for skilled workers and the supply of firm-provided training}.
\newblock \bibinfo{journal}{\emph{Oxford Economic Papers}} \bibinfo{volume}{68}, \bibinfo{number}{1} (\bibinfo{year}{2016}), \bibinfo{pages}{238--257}.
\newblock


\bibitem[Budhwar et~al\mbox{.}(2023)]%
        {budhwar2023human}
\bibfield{author}{\bibinfo{person}{Pawan Budhwar}, \bibinfo{person}{Soumyadeb Chowdhury}, \bibinfo{person}{Geoffrey Wood}, \bibinfo{person}{Herman Aguinis}, \bibinfo{person}{Greg~J Bamber}, \bibinfo{person}{Jose~R Beltran}, \bibinfo{person}{Paul Boselie}, \bibinfo{person}{Fang Lee~Cooke}, \bibinfo{person}{Stephanie Decker}, \bibinfo{person}{Angelo DeNisi}, {et~al\mbox{.}}} \bibinfo{year}{2023}\natexlab{}.
\newblock \showarticletitle{Human resource management in the age of generative artificial intelligence: Perspectives and research directions on ChatGPT}.
\newblock \bibinfo{journal}{\emph{Human Resource Management Journal}} \bibinfo{volume}{33}, \bibinfo{number}{3} (\bibinfo{year}{2023}), \bibinfo{pages}{606--659}.
\newblock


\bibitem[Burgard et~al\mbox{.}(2009)]%
        {burgard2009perceived}
\bibfield{author}{\bibinfo{person}{Sarah~A Burgard}, \bibinfo{person}{Jennie~E Brand}, {and} \bibinfo{person}{James~S House}.} \bibinfo{year}{2009}\natexlab{}.
\newblock \showarticletitle{Perceived job insecurity and worker health in the United States}.
\newblock \bibinfo{journal}{\emph{Social science \& medicine}} \bibinfo{volume}{69}, \bibinfo{number}{5} (\bibinfo{year}{2009}), \bibinfo{pages}{777--785}.
\newblock


\bibitem[Chen et~al\mbox{.}(2024a)]%
        {chen2024landscape}
\bibfield{author}{\bibinfo{person}{Le Chen}, \bibinfo{person}{Nesreen~K Ahmed}, \bibinfo{person}{Akash Dutta}, \bibinfo{person}{Arijit Bhattacharjee}, \bibinfo{person}{Sixing Yu}, \bibinfo{person}{Quazi~Ishtiaque Mahmud}, \bibinfo{person}{Waqwoya Abebe}, \bibinfo{person}{Hung Phan}, \bibinfo{person}{Aishwarya Sarkar}, \bibinfo{person}{Branden Butler}, {et~al\mbox{.}}} \bibinfo{year}{2024}\natexlab{a}.
\newblock \showarticletitle{The Landscape and Challenges of HPC Research and LLMs}.
\newblock \bibinfo{journal}{\emph{CoRR}} (\bibinfo{year}{2024}).
\newblock


\bibitem[Chen et~al\mbox{.}(2024b)]%
        {chen2024ompgpt}
\bibfield{author}{\bibinfo{person}{Le Chen}, \bibinfo{person}{Arijit Bhattacharjee}, \bibinfo{person}{Nesreen Ahmed}, \bibinfo{person}{Niranjan Hasabnis}, \bibinfo{person}{Gal Oren}, \bibinfo{person}{Vy Vo}, {and} \bibinfo{person}{Ali Jannesari}.} \bibinfo{year}{2024}\natexlab{b}.
\newblock \showarticletitle{Ompgpt: A generative pre-trained transformer model for openmp}. In \bibinfo{booktitle}{\emph{European Conference on Parallel Processing}}. Springer, \bibinfo{pages}{121--134}.
\newblock


\bibitem[Chen et~al\mbox{.}(2023a)]%
        {chen2023data}
\bibfield{author}{\bibinfo{person}{Le Chen}, \bibinfo{person}{Xianzhong Ding}, \bibinfo{person}{Murali Emani}, \bibinfo{person}{Tristan Vanderbruggen}, \bibinfo{person}{Pei-Hung Lin}, {and} \bibinfo{person}{Chunhua Liao}.} \bibinfo{year}{2023}\natexlab{a}.
\newblock \showarticletitle{Data race detection using large language models}. In \bibinfo{booktitle}{\emph{Proceedings of the SC'23 Workshops of The International Conference on High Performance Computing, Network, Storage, and Analysis}}. \bibinfo{pages}{215--223}.
\newblock


\bibitem[Chen et~al\mbox{.}(2023b)]%
        {chen2023lm4hpc}
\bibfield{author}{\bibinfo{person}{Le Chen}, \bibinfo{person}{Pei-Hung Lin}, \bibinfo{person}{Tristan Vanderbruggen}, \bibinfo{person}{Chunhua Liao}, \bibinfo{person}{Murali Emani}, {and} \bibinfo{person}{Bronis De~Supinski}.} \bibinfo{year}{2023}\natexlab{b}.
\newblock \showarticletitle{Lm4hpc: Towards effective language model application in high-performance computing}. In \bibinfo{booktitle}{\emph{International Workshop on OpenMP}}. Springer, \bibinfo{pages}{18--33}.
\newblock


\bibitem[Chen et~al\mbox{.}(2022)]%
        {chen2022multi}
\bibfield{author}{\bibinfo{person}{Le Chen}, \bibinfo{person}{Quazi~Ishtiaque Mahmud}, {and} \bibinfo{person}{Ali Jannesari}.} \bibinfo{year}{2022}\natexlab{}.
\newblock \showarticletitle{Multi-view learning for parallelism discovery of sequential programs}. In \bibinfo{booktitle}{\emph{2022 IEEE International Parallel and Distributed Processing Symposium Workshops (IPDPSW)}}. IEEE, \bibinfo{pages}{295--303}.
\newblock


\bibitem[Ding et~al\mbox{.}(2023)]%
        {ding2023hpc}
\bibfield{author}{\bibinfo{person}{Xianzhong Ding}, \bibinfo{person}{Le Chen}, \bibinfo{person}{Murali Emani}, \bibinfo{person}{Chunhua Liao}, \bibinfo{person}{Pei-Hung Lin}, \bibinfo{person}{Tristan Vanderbruggen}, \bibinfo{person}{Zhen Xie}, \bibinfo{person}{Alberto Cerpa}, {and} \bibinfo{person}{Wan Du}.} \bibinfo{year}{2023}\natexlab{}.
\newblock \showarticletitle{Hpc-gpt: Integrating large language model for high-performance computing}. In \bibinfo{booktitle}{\emph{Proceedings of the SC'23 Workshops of The International Conference on High Performance Computing, Network, Storage, and Analysis}}. \bibinfo{pages}{951--960}.
\newblock


\bibitem[Iqbal(2010)]%
        {iqbal2010employee}
\bibfield{author}{\bibinfo{person}{Adnan Iqbal}.} \bibinfo{year}{2010}\natexlab{}.
\newblock \showarticletitle{Employee turnover: Causes, consequences and retention strategies in the Saudi organizations}.
\newblock \bibinfo{journal}{\emph{The Business Review, Cambridge}} \bibinfo{volume}{16}, \bibinfo{number}{2} (\bibinfo{year}{2010}), \bibinfo{pages}{275--281}.
\newblock


\bibitem[Jain et~al\mbox{.}(2020)]%
        {jain2020explaining}
\bibfield{author}{\bibinfo{person}{Praphula~Kumar Jain}, \bibinfo{person}{Madhur Jain}, {and} \bibinfo{person}{Rajendra Pamula}.} \bibinfo{year}{2020}\natexlab{}.
\newblock \showarticletitle{Explaining and predicting employees’ attrition: a machine learning approach}.
\newblock \bibinfo{journal}{\emph{SN Applied Sciences}} \bibinfo{volume}{2}, \bibinfo{number}{4} (\bibinfo{year}{2020}), \bibinfo{pages}{757}.
\newblock


\bibitem[Latha(2013)]%
        {latha2013study}
\bibfield{author}{\bibinfo{person}{K~Lavanya Latha}.} \bibinfo{year}{2013}\natexlab{}.
\newblock \showarticletitle{A study on employee attrition and retention in manufacturing industries}.
\newblock \bibinfo{journal}{\emph{BVIMSR’s Journal of Management Research (BJMR)}} \bibinfo{volume}{5}, \bibinfo{number}{1} (\bibinfo{year}{2013}), \bibinfo{pages}{1--23}.
\newblock


\bibitem[Liga and Robaldo(2023)]%
        {liga2023fine}
\bibfield{author}{\bibinfo{person}{Davide Liga} {and} \bibinfo{person}{Livio Robaldo}.} \bibinfo{year}{2023}\natexlab{}.
\newblock \showarticletitle{Fine-tuning GPT-3 for legal rule classification}.
\newblock \bibinfo{journal}{\emph{Computer Law \& Security Review}}  \bibinfo{volume}{51} (\bibinfo{year}{2023}), \bibinfo{pages}{105864}.
\newblock


\bibitem[Mohan(2024)]%
        {mohan2024management}
\bibfield{author}{\bibinfo{person}{Sai~Krishnan Mohan}.} \bibinfo{year}{2024}\natexlab{}.
\newblock \showarticletitle{Management consulting in the artificial intelligence--LLM era}.
\newblock \bibinfo{journal}{\emph{Management Consulting Journal}} \bibinfo{volume}{7}, \bibinfo{number}{1} (\bibinfo{year}{2024}), \bibinfo{pages}{9--24}.
\newblock


\bibitem[Pallathadka et~al\mbox{.}(2022)]%
        {pallathadka2022attrition}
\bibfield{author}{\bibinfo{person}{Harikumar Pallathadka}, \bibinfo{person}{V~Hari Leela}, \bibinfo{person}{Sushant Patil}, \bibinfo{person}{BH Rashmi}, \bibinfo{person}{Vipin Jain}, {and} \bibinfo{person}{Samrat Ray}.} \bibinfo{year}{2022}\natexlab{}.
\newblock \showarticletitle{Attrition in software companies: Reason and measures}.
\newblock \bibinfo{journal}{\emph{Materials Today: Proceedings}}  \bibinfo{volume}{51} (\bibinfo{year}{2022}), \bibinfo{pages}{528--531}.
\newblock


\bibitem[Raza et~al\mbox{.}(2022)]%
        {raza2022predicting}
\bibfield{author}{\bibinfo{person}{Ali Raza}, \bibinfo{person}{Kashif Munir}, \bibinfo{person}{Mubarak Almutairi}, \bibinfo{person}{Faizan Younas}, {and} \bibinfo{person}{Mian Muhammad~Sadiq Fareed}.} \bibinfo{year}{2022}\natexlab{}.
\newblock \showarticletitle{Predicting employee attrition using machine learning approaches}.
\newblock \bibinfo{journal}{\emph{Applied Sciences}} \bibinfo{volume}{12}, \bibinfo{number}{13} (\bibinfo{year}{2022}), \bibinfo{pages}{6424}.
\newblock


\bibitem[Singh(2023)]%
        {singh2023exploring}
\bibfield{author}{\bibinfo{person}{Vasudha Singh}.} \bibinfo{year}{2023}\natexlab{}.
\newblock \emph{\bibinfo{title}{Exploring the role of large language model (llm)-based chatbots for human resources}}.
\newblock \bibinfo{thesistype}{Ph.\,D. Dissertation}.
\newblock


\bibitem[Team et~al\mbox{.}(2023)]%
        {gemini-paper}
\bibfield{author}{\bibinfo{person}{Gemini Team}, \bibinfo{person}{Rohan Anil}, \bibinfo{person}{Sebastian Borgeaud}, \bibinfo{person}{Yonghui Wu}, \bibinfo{person}{Jean-Baptiste Alayrac}, \bibinfo{person}{Jiahui Yu}, \bibinfo{person}{Radu Soricut}, \bibinfo{person}{Johan Schalkwyk}, \bibinfo{person}{Andrew~M Dai}, \bibinfo{person}{Anja Hauth}, {et~al\mbox{.}}} \bibinfo{year}{2023}\natexlab{}.
\newblock \showarticletitle{Gemini: a family of highly capable multimodal models}.
\newblock \bibinfo{journal}{\emph{arXiv preprint arXiv:2312.11805}} (\bibinfo{year}{2023}).
\newblock


\bibitem[Touvron et~al\mbox{.}(2023)]%
        {llama-paper}
\bibfield{author}{\bibinfo{person}{Hugo Touvron}, \bibinfo{person}{Thibaut Lavril}, \bibinfo{person}{Gautier Izacard}, \bibinfo{person}{Xavier Martinet}, \bibinfo{person}{Marie-Anne Lachaux}, \bibinfo{person}{Timoth{\'e}e Lacroix}, \bibinfo{person}{Baptiste Rozi{\`e}re}, \bibinfo{person}{Naman Goyal}, \bibinfo{person}{Eric Hambro}, \bibinfo{person}{Faisal Azhar}, {et~al\mbox{.}}} \bibinfo{year}{2023}\natexlab{}.
\newblock \showarticletitle{Llama: Open and efficient foundation language models}.
\newblock \bibinfo{journal}{\emph{arXiv preprint arXiv:2302.13971}} (\bibinfo{year}{2023}).
\newblock


\bibitem[Vaswani(2017)]%
        {vaswani2017attention}
\bibfield{author}{\bibinfo{person}{A Vaswani}.} \bibinfo{year}{2017}\natexlab{}.
\newblock \showarticletitle{Attention is all you need}.
\newblock \bibinfo{journal}{\emph{Advances in Neural Information Processing Systems}} (\bibinfo{year}{2017}).
\newblock


\end{thebibliography}


\end{document}